\documentclass[11pt,a4paper]{article}
\usepackage[hyperref]{acl2021}
\usepackage{times}
\usepackage{latexsym}

\usepackage{microtype}

\usepackage{multirow}
\usepackage{graphicx}
\usepackage[tight,footnotesize]{subfigure}
\usepackage{diagbox}

\aclfinalcopy 
\def\aclpaperid{***} 

\title{SimpleMTOD: A Simple Language Model for Multimodal Task-Oriented Dialogue with Symbolic Scene Representation}

\author{Bhathiya Hemanthage , Christian Dondrup, Phil Bartie, Oliver Lemon$\dagger$ \\
    School of Mathematical and Computer Sciences \\ Heriot-Watt University, $\dagger$Alana AI\\
  \{\texttt{hsb2000, c.dondrup, phil.bartie, o.lemon}\}@hw.ac.uk \\
  }

\date{}

\begin{document}
\maketitle
\begin{abstract}
SimpleMTOD is a simple language model which recasts several sub-tasks in
multimodal task-oriented dialogues as sequence prediction tasks. SimpleMTOD is built on a large-scale transformer-based auto-regressive architecture, which has already proven to be successful in uni-modal task-oriented dialogues, and effectively leverages transfer learning from pre-trained GPT-2. In-order to capture the semantics of  visual scenes, we introduce both local and \textit{de-localized} tokens for objects within a scene. De-localized tokens represent the  type of an object rather than the specific object itself and so possess a consistent meaning across the dataset. SimpleMTOD  achieves a state-of-the-art BLEU score (0.327) in the Response Generation sub-task of the SIMMC 2.0 test-std dataset while performing on par in other multimodal sub-tasks: Disambiguation, Coreference Resolution, and Dialog State Tracking.  This is despite taking a minimalist approach for extracting visual (and non-visual) information. In addition the model does not rely on task-specific architectural changes such as classification heads.
\end{abstract}

\section{Introduction}

Multimodal conversational agents have witnessed a rapidly growing level of interest among the conversational AI community as well as within the computer vision community.
Most multimodal conversational datasets to-date are an extension of visual question answering (VQA) \cite{journals/corr/DasKGSYMPB16, conf/cvpr/HudsonM19}. Consequently building upon the success of other visio-linguistic tasks such as VQA, state-of-the-art multimodal conversational agents commonly depend on non-autoregressive models \cite{journals/corr/abs-2004-13278, journals/corr/abs-1912-02379} most of which are based on BERT \cite{devlin2018pretraining}.  

However,  dialogues with such systems significantly differ from what the conversational AI community has typically viewed as a multi-turn dialogue. First, most of the current multimodal dialogue datasets are  focused on querying the visual content whereas {\it external knowledge bases}  have been an integral part of traditional unimodal dialogue datasets \cite{Budzianowski_2018, Galley2019GroundedRG}. Second, in traditional unimodal dialogues, co-reference resolution (explicitly or implicitly) plays a major role within the dialogues. Additionally, state-of-the-art unimodal conversational agents predominantly rely on GPT-based auto-regressive models \cite{radford2018improving}  due to their proven language generation capabilities \cite{journals/corr/abs-2005-05298,hosseiniasl2020simple,conf/acl/HamLJK20}. The SIMMC 2.0 \cite{kottur-etal-2021-simmc} task-oriented dialogue dataset bridges this gap between multimodality and the more traditional view of a multi-turn dialogue. Due to the simultaneous presence of signals from multiple modalities, which a user can refer to at any point in the conversation, the multimodal task-oriented dialogues proposed in the SIMMC 2.0 are  challenging compared to both text-only counterparts and \textit{image querying} dialogue datasets.

In spite of the inherent complexity of multimodal dialogues, we propose SimpleMTOD, recasting all sub-tasks into a simple language model. SimpleMTOD combines the idea of \textit{'de-localized visual object representations'} with a GPT-like auto-regressive architecture. The idea of de-localized representations stems from the analogous process of  \textit{de-lexicalization} that has been extensively used in task-oriented dialogues. In de-lexicalization \citet{conf/acl/MrksicSWTY17}, slot-values such as {\it vegan} are replaced by a more general abstracted token such as {\it food-type}. Likewise, when de-localized, objects are represented by the catalogue type of the object instance rather than the instance itself. These de-localized tokens then possess a consistent meaning throughout the dataset.

 Along with the dataset, \citet{moon2020situated} propose four benchmark tasks decomposing multi-modal task oriented dialogue into sub-tasks: Multimodal Disambiguation, Multimodal Co-reference Resolution,  Multimodal Dialog State Tracking, and Response Generation. The first three tasks deal with the dialogue context understanding, analogous to NLU and DST in unimodal agents. The last task is similar to unimodal NLG, but expects the generated responses to be sensible within a multimodal context with visual signals and associated knowledge base. 

The main objective this work is to evaluate the effectiveness of de-localized object representations within SimpleMTOD. Despite the simplicity, SimpleMTOD achieves the state-of-the-art BLEU score of 0.327 for assistant response generation in the SIMMC2.0 test-std \footnote{The testing dataset (test-std) is not publicly available and was part of the SIMMC 2.0 challenge used for scoring the submitted systems.} dataset . Furthermore, the model achieves an accuracy of 93.6\% in Multimodal Disambiguation (MM-Disambiguation),  Object-F1 of 68.1\% in Multimodal Co-reference Resolution (MM-Coref), and  87.7\% (Slot-F1) and 95.8 (Intent-F1) in Multimodal Dialogue State Tracking (MM-DST). Other than the proposed benchmark settings, we also evaluate SimpleMTOD in an end-to-end setting. Major contributions of our work are as follows:

\begin{itemize}
    \item We formalise notion of \textit{multimodal task oriented dialogues }as an end-to-end task. 
    \item We propose a GPT-based simple language model combined with visual object de-localization and token based spatial information representation, that addresses four sub-tasks in multimodal dialogue state tracking with a {\it single architecture}.
    \item We analyse the behaviour of our model using salience scores from  the Ecco \citep{alammar-2021-ecco} framework, which provide an intuition into which previous token mostly influence predicting the next token.
\end{itemize}

\section{Background}

\begin{figure*}
    \centering
    \subfigure[]{\includegraphics[width=1.5\columnwidth]{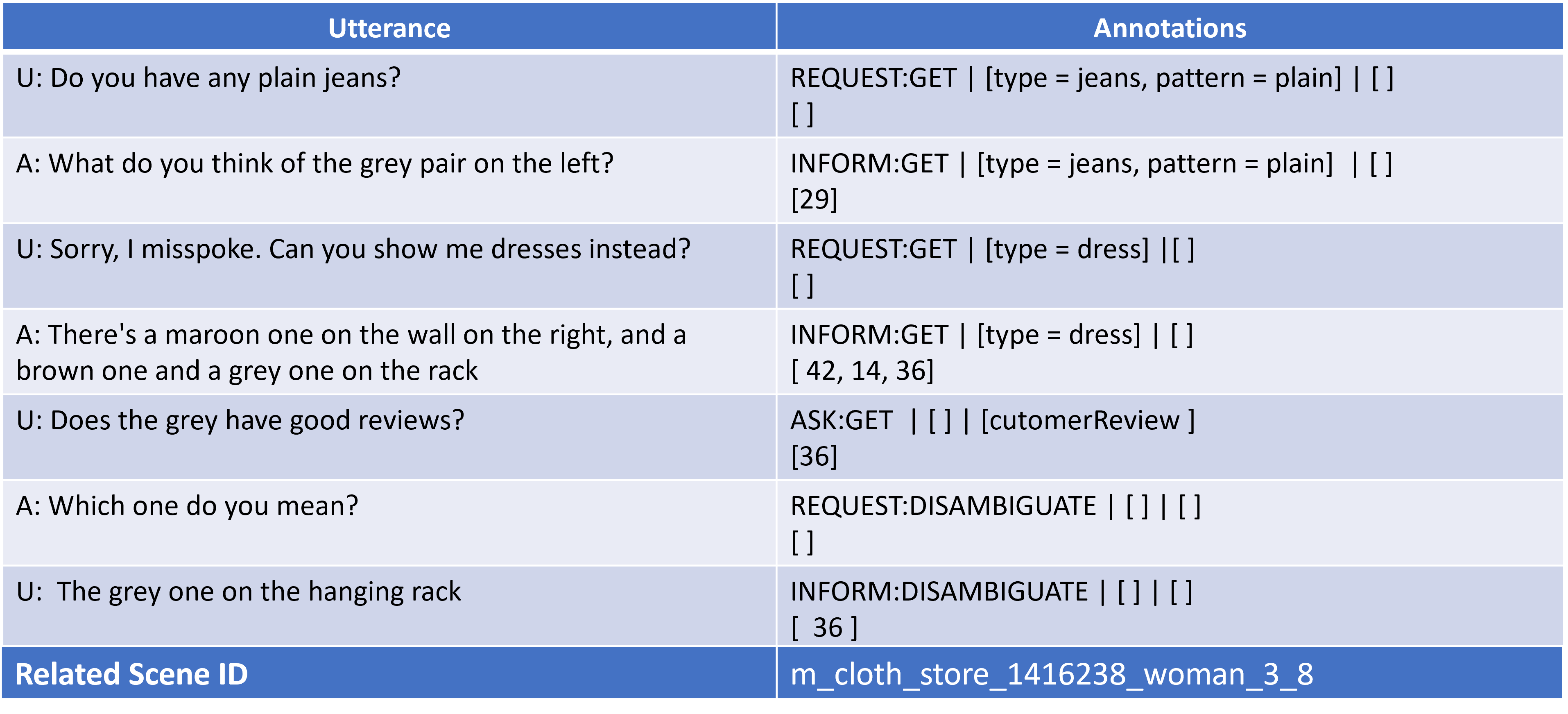}\label{sugfig:app_dial}}
    \subfigure[]{ \includegraphics[width=1.5\columnwidth]{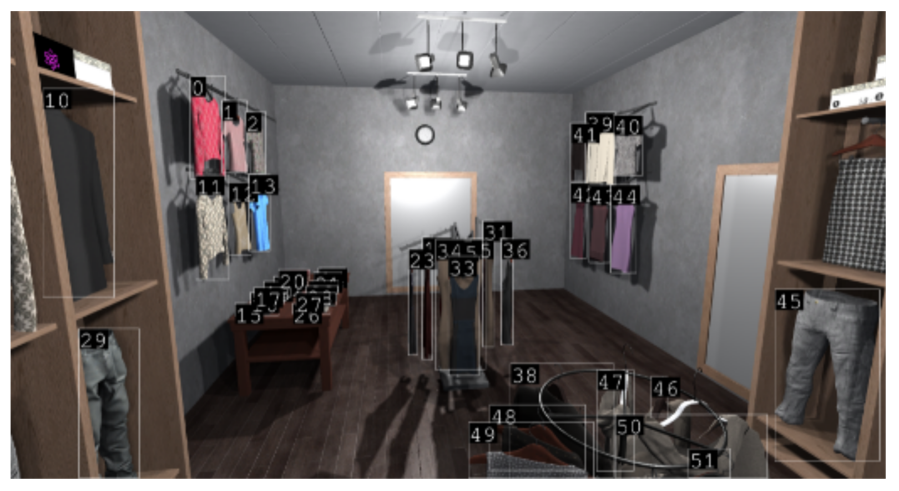}\label{sugfig:app_image}}
    \subfigure[]{ \includegraphics[width=1.5\columnwidth]{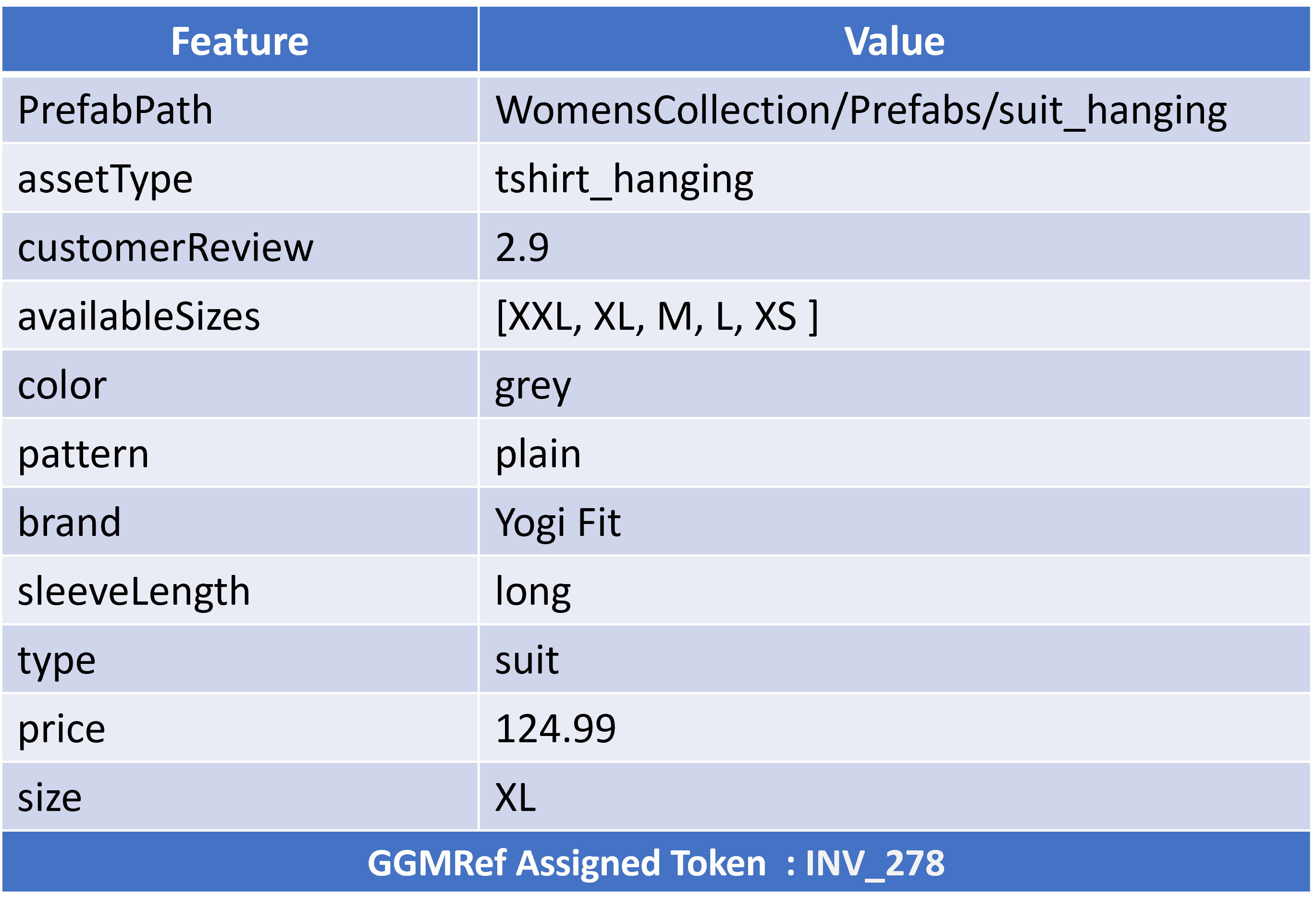}\label{sugfig:app_inv}}
    \caption{ Sample dialogue instance in SIMMC 2.0: a) First four turns of a sample dialogue with user and system transcript annotations. U: and A: tokens are used to differentiate user and system utterances respectively. First row of annotations are in INTENT $|$  SLOT-VALUE $|$  REQUEST-SLOTS  format. Second row identifies referred canonical objects id tags in the utterance (e.g.\ [29]). It should be noted that, these object ids are specific to a given scene. In the case of user utterances, this identifier is the target of the MM-Coref task. b) Sample image with cannonical object id tags over items. This image is mapped to the dialogue by scene id. c) Single entry of the fashion object meta-data file. }
    \label{fig:sample_dialogue}
\end{figure*}

Traditional task-oriented dialogue datasets consist of a dialogue corpus, a dialogue ontology with a pre-defined set of slot-value pairs, and annotations required for related sub-tasks in a set of domains \cite{Budzianowski_2018}.
The SIMMC 2.0 dataset follows a similar structure and contains dialogues in both the fashion and  the furniture domains. However, in the SIMMC 2.0 multimodal dialogue corpus, each dialogue is also associated with an image representing the scene where each dialogue takes place. A \textit{scene} is made by re-arranging a known set of items (objects) in different configurations. Along with the raw-image, the dataset provides a file (scene JSON) containing details of the images such as objects and relationships between objects. Furthermore, a meta-data file  contains visual and non-visual attributes of objects that recur within a scene.

\subsection{Benchmark Tasks}
\paragraph{Multimodal Disambiguation:} In real-world conversations, references made by humans related to objects or entities can be ambiguous. For example, consider \textit{A: Blue trousers are priced at \$149.99. U: What about the red ones?}, in a setting where there are multiple red trousers. In these situations, there is insufficient information available for co-reference resolution. This task is aimed at identifying such ambiguous scenarios, given the dialogue history. 

\paragraph{Multimodal Co-reference Resolution:}
The goal of this task is to resolve any reference in a user utterance to canonical object ids of the object as defined per each scene (see image in Figure \ref{sugfig:app_image}). Users may refer to 1) dialogue context 2) visual context, or 3) both.

\paragraph{Mutltimodal Dialogue State Tracking:}
Similar to unimodal DST, this tracks the belief states of users across multiple turns. The belief state consists of an intent, slot-value pairs, and user requested slots.

\paragraph{Assistant Response Generation}
Given the user utterance, ground-truth APIs, and ground-truth cannonical object ids (with meta-data), the model needs to generate a natural language response describing objects as \textit{observed and understood} by the user.

\section{Methods}

\begin{figure*}[t!]
    \centering
    \includegraphics[width=2.0\columnwidth]{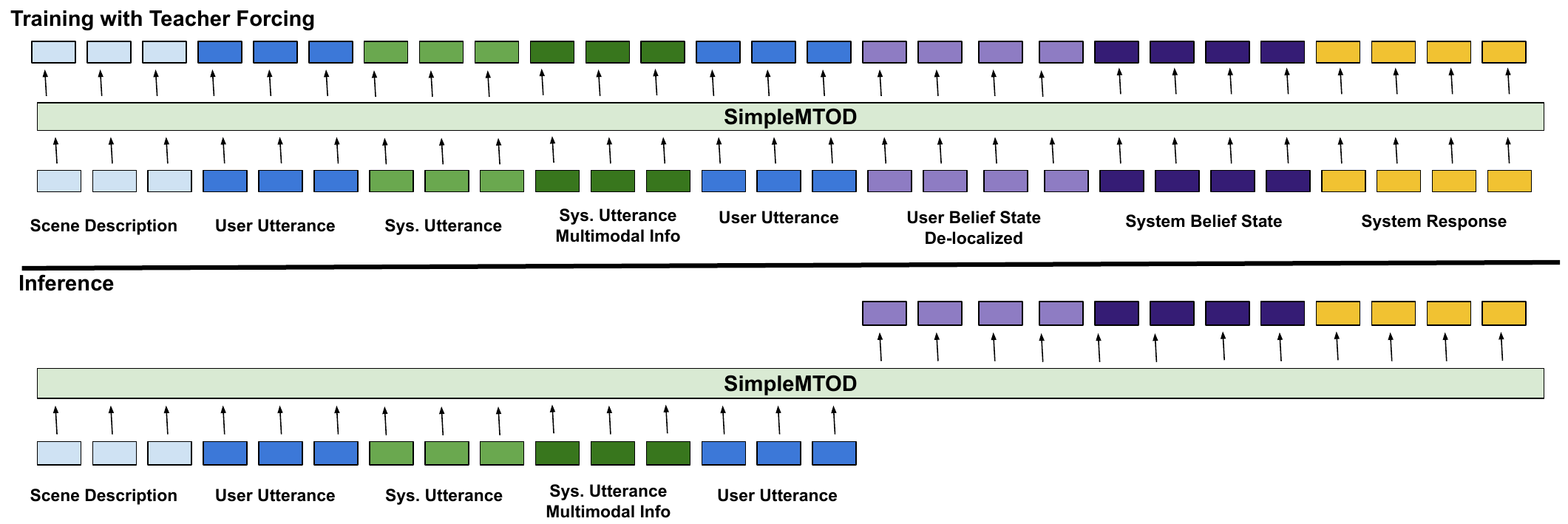}
    \caption{SimpleMTOD architecture with training and inference time setting}
    \label{fig:mtod_arch}
\end{figure*}

In the first part of this section, we model multimodal task oriented dialogues as a sequence generation task. We define the problem in a more general setup and discuss some empirical limitations applied to the model.

\subsection{Multimodal Task-Oriented Dialogues}

\label{MM_TOD}
Similar to unimodal setting, we view dialogue state (belief-state) tracking, action prediction, and response generation to be  the core components of  multi-modal task-oriented dialogues. However, outputs of each of the sub-tasks should be conditioned not only on the dialogue history, but also on the associated scene.

Multimodal dialogues consist of multiple turns. In a turn \(t\), there exists an associated  visual scene   \(V_t\), the user-provided input \(U_t\) and the system-generated  response \(S_t\). Theoretically, the dialogue context can be denoted as \(C_t = [V_0,U_0, S_0|V_0, . . . S_{t-1}|M_{t-1},V_t, U_t]\). Here \(S_{t-1}|M_{t-1}\) denotes that the statement \(S_{t-1}\) is associated with the representation of multimodal information such as objects viewed and mentioned to the user during that turn.

Given the context, \(C_t\), SimpleMTOD generates the belief-state \(B_t\):
\begin{equation}
    B_t = SimpleMTOD(C_t)
\label{eq_belief}
\end{equation}

\(B_t\) is a concatenation of intent, slot-values, requested slots, and resolved object references \(\\MRef_t\).

However, it should be noted that, SimpleMTOD models the context as  \(C_t = [V_t, U_{t-n}, S_{t-n}|M_{t-n}, . . . S_{t-1}|M_{t-1},U_t, ]\) where the $n$ is the context window. Major deviations from the theoretical representation of \(C_t\) are, 1) we ignore the history of visual signals and only consider the current visual scene; 2) we consider only \(n\) previous turns in contrast to the entire dialogue.

Then, in a more generalized setting where the system have access to an external database, which can be queried,\(B_t\) would be used to retrieve database results \(D_t\). These \(D_t\) along with context and belief states can be  used to generate the system action \(A_t\).

\begin{equation}
    A_t = SimpleMTOD(C_t, B_t, D_t)
\label{eq_action}
\end{equation}

Action \(A_t\) is a triplet containing system intent, slot-value pairs, and details on requested slots. However, in our setup, no such database exists. Hence we model action \(A_t\) from \(B_t\) and \(C_t\) keeping \(D_t=\emptyset\).

Finally, the concatenation of the context, belief state,  (database results), and action is used to generate system responses \(S_t\).
\begin{equation}
    S_t = SimpleMTOD(C_t, B_t, D_t, A_t)
\end{equation}

\subsection{De-localized Visual Representation}
\label{sec:de-local}

\begin{figure}
    \centering
    \includegraphics[width=\columnwidth]{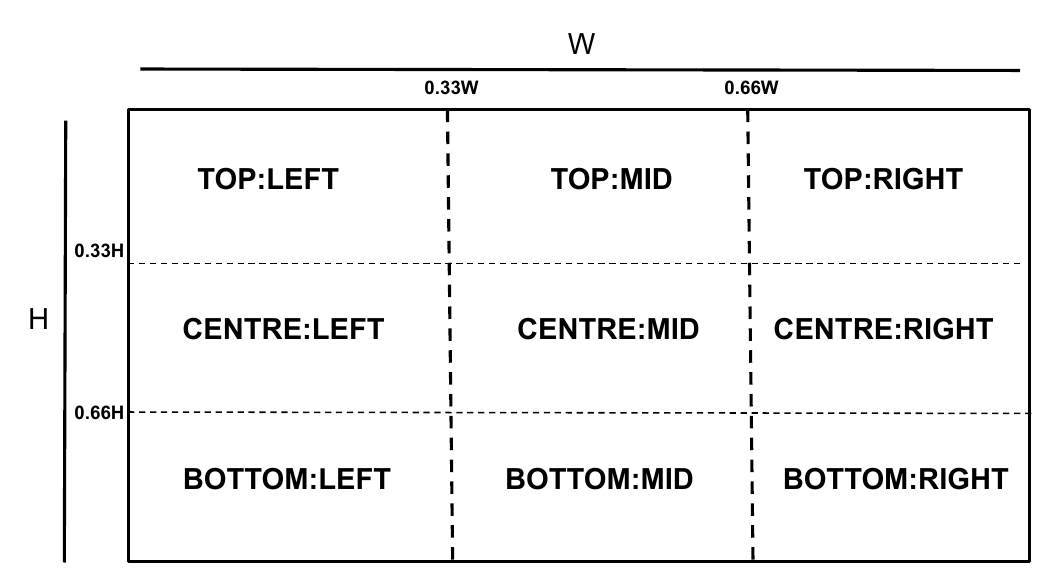}
    \caption{A scene is divided into 9 regions. Each region is identified by combination of 2 tokens.}
    \label{fig:regions}
\end{figure}

Here we discuss how visual information of a scene is represented within the SimpleMTOD as de-localized tokens and how \(V_t\) is derived from those tokens. 

In the SIMMC 2.0 dataset a scene is  a spatial configuration of a set of object instances. From  here on we will refer to these instances simply as objects. Probable types of these objects are pre-defined in two meta-data files, with one for each domain. We will refer to these files as catalogues and an entry of these catalogues as a catalogue-item. See Figure\ref{sugfig:app_inv} for an example catalogue-item with visual and non-visual attributes defined. For benchmark tasks, non-visual attributes can be used during inference while visual attributes are not allowed. However, we use neither of these attributes in the SimpleMTOD visual representation explained below.

In our setup, we assign a unique token (eg: \textit{INV\_278}) to each catalogue-item. These catalogue-items are used as a de-localized version of objects within a scene. While these catalogue-item tokens are consistent across the entire dataset, spatial relationships associated with the objects will be lost. Therefore we encode spatial details of objects as follows:
Each scene is divided into 9 regions as shown in Figure \ref{fig:regions}. Every object is assigned to a region based on the center-point of the object bounding box. Then concatenation of catalogue-item tokens and assigned region description (eg: \textit{INV\_278@TOP:LEFT}) tokens are used as object representations. A scene-description is obtained by concatenating all such tokens representing every object within a scene. This is our \(V_t\) in SimpleMTOD.

\subsection{SimpleMTOD Training and Inference}
 
 For training, we follow routine causal language modeling with teacher forcing.  A training sequence \(X_t\) in SimpleMTOD is obtained by concatenating all the components; context,user belief state, database results (which is null in our case), system actions and system utterance.
 \begin{equation}
     X_t = [C_t, B_t, D_t, A_t, S_t]
 \end{equation}
 
In terms of tokens, \(X_t\) can be denoted as \( X_t = (x^0_t,x^1_t,....x^{n(t)}_t)\) when \(n(t)\) represent the number of tokens in turn \(t\). In general, the goal of the model is to learn \(\rho(X)\) given \( X = (x^0,x^1,..x^i..x^n)\) :

 \begin{equation}
    \rho(X)  = \Pi_{i=1}^{n}\rho(x^i|x^{<i})
 \end{equation}
 
 For this, we train the neural network with parameterization \(\theta\) minimizing the negative log-likelihood over the multimodal dialogue corpus \(MD\) where \( MD=\{X_1,X_2....X_{|MD\|}\} \) . However, in our setup the tokens related to scene-description \(V\) are ignored during the loss calculation. When \(n(V)\) is the number of tokens related to the scene description:
 
 \begin{equation}
     L(D) = -\sum_{t=1}^{|MD|}{\sum_{i={n(V)}}^{n(t)} {log\rho_{\theta}(x^i_t|x^{<i}_t)}}
 \end{equation}
 
 During inference, the learnt parameter \(\theta\) is used to predict a token at a time. Unlike training time where ground-truth tokens are used  every time, generated tokens become part of the left-context. For inference, we stick to a simple greedy prediction approach with top-k=1. That is we always generate the token with highest probability as the next token.

\section{Experiments}

\begin{table*}[h!]
\centering
\begin{tabular}{l c c c c}
    \hline
    Model & Intent-F1 &  Slot-F1 & Request Slot-F1 & Joint Accuracy  \\
    \hline
    GPT-2 Baseline &  94.5 & 81.7 & 89.6 & 44.6 \\
    MTN-SIMMC &  94.3 & 74.8 & 85.4 & 28.3 \\
    SimpleMTOD$_{Sub}$ & \textbf{95.8} & 83.3 & 89.7 & 57.3 \\
    SimpleMTOD & 94.0 & \textbf{85.8} & \textbf{91.7} & \textbf{63.1} \\
    \hline
\end{tabular}
\caption{Evaluation results for MM-DST task on Devtest split}
\label{table:mm-dst}
\end{table*}

In Section \ref{MM_TOD} we defined an end-to-end setting for SimpleMTOD. However, some of the benchmark tasks allow more ground-truth information to be utilized during training and inference time.

For the MM-Disambiguation task, we consider two setups. In the task-specific scenario, we train the model to predict YES or NO tokens directly from context \(C_t\). In the end-to-end setup, we consider the label to be YES only if the system intent predicted is to Disambiguate. Two similar setups are considered for MM-Coref as well. It should be noted that end-to-end version of SimpleMTOD predicts de-localized tokens with spatial information and we obtain the canonical object id by reversing the de-localization process explained in Section \ref{sec:de-local}. If multiple objects were found in the same region with same catalogue-item token, the area of the object bounding box is used as a tie-breaker. In the case of assistant response generation, the benchmark task defined in SIMMC 2.0 allows ground-truth system belief state to be used as an input. Therefore, we evaluate both from action response generation as well as end-to-end setting.

\subsection{Baselines}
We consider 2 baselines which were provided as part of the SIMMC2.0 challenege.
\paragraph{GPT-2:} This  extends \citet{conf/acl/HamLJK20} to multi modal task-oriented dialogues, encoding objects in a scene using canonical object ids concatenated with the token OBJECT\_ID. For the MM-Disambiguation task, a classification head is used, while other tasks are modeled in a generative manner.
\paragraph{Multimodal Transformer Networks (MTN):} Adapts \citet{conf/acl/LeSCH19} (only) for the MM-DST and Response Generation sub-tasks   \footnote{MTN-SIMMC2 implementation \url{https://github.com/henryhungle/MTN/tree/simmc2}}. In contrast to the auto-regressive modeling of SimpleMTOD, MTN uses an encoder-decoder architecture.

\subsection{Training and Evaluation}
We follow the experimental setup of the  SIMMC 2.0 challenge with same dataset-splits, inference time limitations, and performance metrics. See Appendix:\ref{sec_train_eval} for details. It should be noted that the test-std split of the SIMMC2.0 dataset is not publicly available and is a held-out set for evaluating submissions to SIMMC2.0 challenge. Therefore, the final version of our model could only be evaluated  on the dev-test split. However, the prior version of the model SimpleMTOD$_{Sub}$, which did not encode region information or scene information, was submitted to the SIMMC2.0 challenge.

\section{Results}

\begin{table}[h!]
\centering
\begin{tabular}{l c c }
    \hline
    Model & Accuracy & Object-F1 \\
    \hline
    GPT-2 Baseline &  73.5 & 36.6\\
    SimpleMTOD$_{Sub}$ & \textbf{92.17} & 67.6\\
    SimpleMTOD & 92.12 & \textbf{73.5}\\
    \hline
\end{tabular}
\caption{Accuracy and Object-F1  scores for MM-Disambiguation and MM-Coref tasks on Devtest split.}
\label{tbl:MM-Dis}
\end{table}

\begin{table}[h!]
\centering
\begin{tabular}{l c}
    \hline
    Model & BLEU  \\
    \hline
    GPT-2 Baseline & 0.192 \\
    MTN-SIMMC & 0.217 \\
    SimpleMTOD$_{Sub}$ & 0.43 \\
    SimpleMTOD(ground truth actions) & \textbf{0.49}  \\
    SimpleMTOD & 0.45\\
    \hline
\end{tabular}
\caption{BLEU scores for Assistant Response Generation task on Devtest split.}
\label{table:asr}
\end{table}

\begin{table*}[h!]
    \centering
    \begin{tabular}{l c c c c c c c c c c c c}
\cline{1-6}
{\multirow{2}{*}{\textbf{Model}}} & \textbf{MM-Disam'n} & \textbf{MM-Coref}  &\multicolumn{2}{c}{\textbf{DST}} & \textbf{Response Generation}\\ \cline{2-6}

{\multirow{2}{*}{}}  & \textbf{Accuracy} & \textbf{Object-F1} & \textbf{Intent-F1} & \textbf{Slot-F1} & \textbf{BLEU} \\ 
\cline{1-6}

{\multirow{1}{*}{\textbf{GPT-2 Baseline}}}  & 73.5 & 44.1 & 94.1 & 83.8 & 0.202 \\ 
{\multirow{1}{*}{\textbf{MTN - Baseline}}}  & NA & NA & 92.8 & 76.7 & 0.211 \\ 
{\multirow{1}{*}{\textbf{Team-2}}}  & NA & \textbf{78.3} & 96.3 & 88.4  & NA  \\ 
{\multirow{1}{*}{\textbf{Team-5}}}  & 93.8 & 56.4 & \textbf{96.4} & 89.3 & 0.295 \\ 
{\multirow{1}{*}{\textbf{Team-6}}}  & \textbf{94.7} & 59.5 & 96.0 & \textbf{91.5}  & 0.322 \\ 
{\multirow{1}{*}{\textbf{SimpleMTOD$_{Sub}$}}}  & 93.6 & 68.2 &  95.8 & 87.7 & \textbf{0.327} \\ \cline{1-6}
\end{tabular}
\caption{Test-std results for SIMMC2.0 Challenge. NA denotes model is not applicable to the particular sub-task. Test-std split of SIMMC2.0 dataset is held-out set, which is not publicly available and used to evaluate submissions in SIMMC2.0 challenge. An earlier version of the system,   SimpleMTOD$_{Sub}$, without scene information,  was submitted for the evaluation.  }
\label{tbl:test-std_results}
\end{table*}

\paragraph{MM-Disambiguation} 
As shown in Table \ref{tbl:MM-Dis} and Column 2 of Table \ref{tbl:test-std_results}, SimpleMTOD$_{Sub}$ achieves accuracy scores of 92.17\% and 93.6 on devtest and test-std respectively when trained to predict YES/NO tokens.  This is a 27\% relative improvement over the GPT-2 based baseline with a classification head. Furthermore, we evaluate the model on the MM-Disambiguation task as part of the end-to-end model. based on the system intent predicted by the model. Here, we consider any \textit{INFORM:DISAMBIGUATE} prediction as a YES. This approach demonstrates a very similar accuracy score of 92.12. The best performing model (94.5\% : Team-6) on test-std, ensembles two models trained on RoBERTa and BART \footnote{This is based on the description provided at: \url{ https://github.com/NLPlab-skku/DSTC10\_SIMMC2.0}}.

\paragraph{ MM-Coref}
Table \ref{tbl:MM-Dis} and the Third column of the Table \ref{tbl:test-std_results} show the MM-Coref Object-F1 scores of on devtest and test-std respectively. SimpleMTOD achieved 68.2 (54\%  relative gain over baseline) in test-std dataset and 67.6 (84\% gain) on the devtest split. While there is no information available on Team-2's leading solution, the BART-based model of Team-4 which is trained end-to-end with task-specific heads achieves 75.8\% on this task.

\paragraph{MM-DST} 
Despite being a simple language model, both our Intent-F1 (95.8\%) and Slot-F1 (87.7\%) scores on test-std split are comparable with complex visual-language models. Furthermore, as in Table \ref{table:mm-dst}, there is significant improvement in the Joint Accuracy scores from 57.3\% to 63.1\% when positional information is used.

\paragraph{Response Generation}
A prior version of the model, SimpleMTOD$_{Sub}$ achieves a state-of-the-art BLEU score of 0.327 on the test-std split of the SIMMC2.0 dataset. This is in comparison with models which rely on sophisticated feature extraction processes. In our view, the simplified representation of visual information preserves and complements the generative capabilities of pre-trained models.  Furthermore, as shown in Table \ref{table:asr}, SimpleMTOD achieves a BLEU score of 0.49 on devtest when the ground-truth actions are used. The end-to-end  version of SimpleMTOD also achieves a BLEU score of 0.45. It should be noted that this is an improvement over the \(SimpleMTOD_{Sub}\) model score of 0.43. This indicates the importance of associating region related information.

\section{Discussion} \label{sec:discussion}
\begin{figure*}[t!]
    \centering
    \includegraphics[width=1.7\columnwidth]{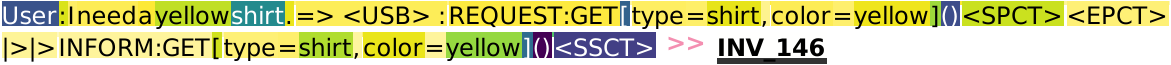}
    \caption{Salience score heat-map when predicting the token \textit{INV\_146} for utterance \textit{I need a yellow shirt} without scene information. Darker colors represents higher salience score. See Figure:\ref{fig:ecco_coref_woscene_deteailed} in appendix for actual values}
    \label{fig:ecco_coref_woscene}
\end{figure*}

\begin{figure*}[t!]
    \captionsetup[subfigure]{labelformat=empty}
    \centering
    \includegraphics[width=1.7\columnwidth]{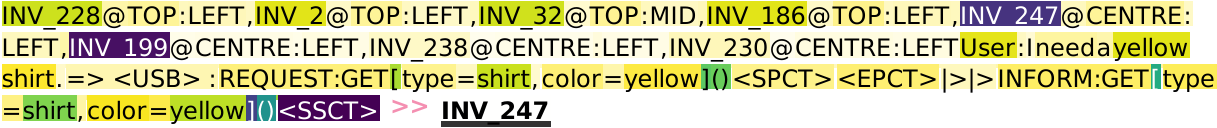}\label{sugfig:247_colormap}
    \caption{Salience scores heat-map  {\it with scene information} when predicting the token \textit{INV\_247} in utterance \textit{I need a yellow shirt}. See Figure:\ref{fig:ecco_coref_scene_detailed} in appendix for actual values}
\label{fig:ecco_coref_scene}
\end{figure*} 

\begin{figure*}[t!]
    \centering
    \includegraphics[width=1.7\columnwidth]{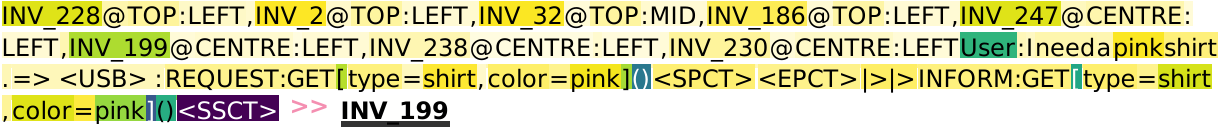}
    \caption{Salience score heat-map when predicting the token \textit{INV\_199}   for modified utterance \textit{I need a pink shirt} See Figure:\ref{fig:ecco_coref_mod_detailed} in appendix for actual values }
    \label{fig:ecco_coref_mod}
\end{figure*}

\begin{table}[h!]
\centering
\begin{tabular}{|l| c c c| }
    \hline
    \diagbox[width=5em]{Feature}{Token} & INV\_146 &  INV\_199 & INV\_247 \\
    \hline
    Color &  yellow & pink & yellow \\
    Type &  shirt & shirt & shirt \\
    \hline
\end{tabular}
\caption{Relevant catalogue items represented by tokens INV\_146, INV\_199, INV\_247. None of these metadata were explicitly presented to the model.}
\label{table:token_meta}
\end{table}

\begin{table*}[h!]
\centering
\begin{tabular}{l c c c }
    \hline
    Original(color=yellow) &  INV\_247 (92.63) & INV\_199 (7.17) & INV\_155(0.08) \\
    Original w/o desc. &  INV\_146(13.75) & INV\_247 (13.04) & INV\_249 (12.60) \\
    Modified(color=pink) &  INV\_199(99.79) & INV\_247 (0.19) & INV\_235($<$0.01) \\
    \hline
\end{tabular}
\caption{ For the example utterances discussed, we inspected top-3 tokens and their confidence scores.}
\label{table:token_confidence}
\end{table*}

In order to understand the behaviour of   SimpleMToD, we use gradient-based salience \citep{conf/emnlp/AtanasovaSLA20} provided with the Ecco framework \citep{alammar-2021-ecco}. Using Ecco, we inspect salience scores for all the tokens in the left side of the token of interest. In the heat-maps presented in this section, darker colors  mean a higher salience score. It should  also be noted that the model assigns high salience scores on separator tokens (such as \textit{$<USB>$, [ , ] }) that define the structure of the generation. While proper attention to the structure is of paramount importance, our \textbf{ discussion   focuses on salience scores assigned to the rest of the tokens, which represent the semantics} of the multimodal conversations.

\paragraph{Effect of De-localization and Scene Descriptions:}
The introduction of de-localized tokens significantly improves the  Object-F1 of MM-coref and joint accuracy of MM-DST. Accordingly, we first analyse the behaviour of the model when predicting co-references. Figures \ref{fig:ecco_coref_scene} and \ref{fig:ecco_coref_woscene} show example utterances with and without scene descriptions respectively. 
In the case where scene description is not provided, the model puts a high salience on tokens `yellow' and `shirt', and predicts the token INV\_146 which represents a yellow color shirt as shown in Table \ref{table:token_meta}. (It should be noted that none of the metadata shown in the diagram are provided to the model explicitly and the model figures this out from globally consistent use of tokens).  However, in this case, a particular catalogue item INV\_146 is not present in the scene. When we observe the confidence values of the prediction from the last layer (shown in Table \ref{table:token_confidence}), it can be seen that the model is not quite certain about the prediction with 13.75 for INV\_146 and 13.04 for INV\_247, both of which represent yellow shirts. This is to indicate that even though the model has learnt to associate object attributes necessary for co-reference resolution, it lacks information to be certain about the prediction. To this end, we provide the model with a scene description as described in \ref{sec:de-local}.  When the scene descriptions are provided, SimpleMTOD  correctly predicts the token INV\_247 with 92.63\% confidence and high salience score over the same token from the scene description, as well as tokens `shirt' and `yellow'.

Additionally from Figure \ref{fig:ecco_coref_scene} it can be noted that INV\_199 also shows a high salience score. From the metadata, we can see it is a pink color shirt. However, there is a significant salience score over the token `yellow' that results in generating the correct token INV\_247 over INV\_199 (which is the second ranked token with only had 7.17 confidence). Extending the analysis, we modified  the original utterance  to \textit{``I need a pink shirt"} and generated the next token, and SimpleMToD accordingly predicted the token INV\_199 (with high confidence of 99.79\%)  as observed in Figure \ref{fig:ecco_coref_mod}.

\paragraph{Effect on Intent prediction:} 

\begin{figure*}[t!]
    \centering
    \includegraphics[width=1.7\columnwidth]{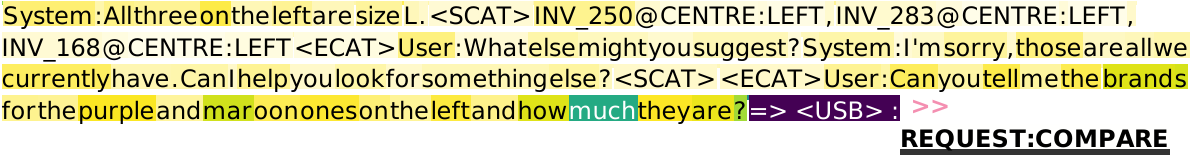}
    \caption{Salience score heat-map when predicting the correct intent token \textit{REQUEST:COMPARE} for the dialogue turn with final utterance \textit{``Can you tell me the brands for the purple and maroon ones on the left and how much they are?"} without providing scene information}
    \label{fig:ecco_intent_woscene}
\end{figure*}

Even though scene descriptions play a key role in overall belief tracking as described earlier, the Intent-F1 score drops from 95.8\% to 94.0\%  when the scene descriptions are encoded. In order to understand the effect, we inspect salience scores when predicting the user intent. It can be observed that when the scene descriptions are omitted, higher salience scores are assigned to the user utterance suggesting more focus on that. However, when the scene information is included, salience scores assigned to the utterance decreased to an extent, resulting in wrong predictions in certain cases. This is to indicate that scene descriptions are either redundant or act as a distractor when we consider intent-detection, which explains reduction in score. Furthermore, this behaviour aligns with our intuition that the intent parts of the user utterances are predominantly language-driven.
Figure \ref{fig:ecco_intent_woscene} shows an example where omitting the scene information produces the correct intent of  \textit{REQUEST:COMPARE}, whereas our final version of SimpleMTOD wrongly predicted the intent as \textit{ASK:GET}

\section{Related Work}

\citet{journals/corr/abs-2005-05298,hosseiniasl2020simple,conf/acl/HamLJK20} are closely related to our work as they all model task-oriented dialogues in an end-to-end manner with  GPT-2-like large-scale transformer-based architectures. However, all those models focus on {\it text-only} task-oriented dialogues. The GPT-2 adaptation \citep{kottur-etal-2021-simmc}, which is provided as a baseline along with the SIMMC2.0 dataset,  is also closely related to our work. However, this baseline   represents visual objects  by canonical ids 
and demonstrates subpar results to our model in all four tasks.

Generative encoder-decoder models \citep{conf/aaai/LiangTCY20,conf/sigdial/ZhaoLLE17}  are a promising alternative to decoder-only (GPT-2 based) dialogue models that have been extensively investigated in unimodal task-oriented dialogues. The MTN-baseline \citep{conf/acl/LeSCH19},  which we compare to, is based on the encoder-decoder architecture. While being inferior with respect to performance in both the tasks considered, this model involves sophisticated feature extraction process.

\citet{conf/acl/MrksicSWTY17} coined the term  `de-lexicalization' for abstraction in neural dialogue state tracking tasks. This idea has been extensively used in goal oriented dialogues. Our notion of de-localized object representation is influenced by this work.

\section{Conclusion}
We explore a simple, single generative architecture  (SimpleMTOD) for several sub-tasks in multimodal task-oriented dialogues.  We build on large-scale auto-regressive transformer-based language modeling, which has been effectively utilized in task-oriented dialogues, and formalize the multimodal task-oriented dialogue as a sequence prediction task.  Our model employs a  `de-localization' mechanism for visual object representation that ensures the consistency of those tokens throughout the dataset. Furthermore, we encoded spatial information of object instances with a very small number of special (globally consistent) tokens. Despite the simplicity in representing  visual information, our model demonstrates comparable or better performance with models that heavily rely on visual feature extraction, on four multimodal sub-tasks in the SIMMC2.0 challenge.

\section{Future Directions}
Most current vision-language research relies on fusing pixel-level vision information with token-level language representations. However, their applicability for dialogues where the language is sophisticated remain sparsely studied. In contrast, we explore a symbolic approach for representing visual information and combining it with auto-regressive language models. While we rely on smaller scale models (with 17 million parameters), our work is readily extendable for large language models (LLMs). Unlike pixel level visual representations, special tokens representing visual information being more similar to the word tokens which the LLMs area trained on, symbolic visual representation would facilitate effective transfer learning.

SimpleMTOD represents visual information using carefully designed input tokens. Capturing these information through semantic scene-graphs, which would provide richer representation, and fusing them with LLMs would be an interesting future direction of research for multimodal dialogues. Development in knowledge-graph based language grounding would complement this line of work.

\section*{Acknowledgements}
This work is partially supported by the European Commission under the Horizon 2020 framework programme for Research and Innovation (H2020-ICT-2019-2, GA no.\ 871245), SPRING project,  https://spring-h2020.eu 

\bibliographystyle{acl_natbib}
\bibliography{anthology,acl2021}
\newpage
\appendix

\section{SIMMC 2.0 Dataset}

The SIMMC 2.0 dataset ( released under CC-BY-NC-SA-4.0 licence) \footnote{https://github.com/facebookresearch/simmc2 } consists of three major components: 
\begin{itemize}
    \item  \textbf{Dialogue Data: }Includes system and user utterance with relevant annotations. Figure \ref{sugfig:app_dial} provide first 4 turns of a sample dialogue.
    \item \textbf{Scene Data:} Set of scenes representing environments in which dialogues take place. Figure \ref{sugfig:app_image} provide the scene related to the dialogue segment shown in Figure \ref{sugfig:app_dial}. Other than raw-images , an json file associated with each image provides detail of objects, such as bounding boxes and spatial relationships (left of, right of, over, under) among objects.
    \item \textbf{Meta-data:} acts as a catalogue of items related to the dialogue corpus. Scene images are made-up by positioning instances of catalogue items in different configurations. Entries contain both visual and non-visual attributes of each item. Visual attributes of items from the meta-data file are not allowed to be used during inference.Figure \ref{sugfig:app_inv} shows a single entry in meta-data file.
\end{itemize}

\subsection{Data Statistics}
\begin{table}[h!]
\centering
\resizebox{0.8\columnwidth}{!}{
\begin{tabular}{l c }
    \hline
    Split & \# of Dialogues\\
    \hline
    Train (64\%) & 7307 \\
    Dev (5\%) & 563\\
    Test-Dev(15\%)	& 1687 \\
    Test-Std (15\%)	& 1687 \\
    \hline
\end{tabular}
}
\caption{Number of dialogues in each split. }
\vspace*{-6mm}
\label{table:teststd}
\end{table}

\section{Training and Evaluation}

 \begin{table}[hbt!]
\begin{tabular}{|l|l|}
\cline{1-2}
\textbf{Task} & \textbf{Metric} \\  \cline{1-2}
{\multirow{1}{*}{MM-Disambiguation}} & Accuracy\\ \cline{1-2}
{\multirow{1}{*}{MM-Coref}} & Object-F1 \\ \cline{1-2} 
{\multirow{2}{*}{MM-DST}} & Intent-F1 \\ \cline{2-2}
{\multirow{2}{*}{}} & Slot-F1 \\  \cline{1-2}
{\multirow{1}{*}{DST+Coref}} & Joint Accuracy \\ \cline{1-2}
{\multirow{1}{*}{Response Generation}} & BLEU-4 \\\cline{1-2}
\end{tabular}
\caption{Evaluation metrics used for different tasks in SIMMC 2.0}
\label{tbl:eval_met}
\end{table}

\label{sec_train_eval}
 We conduct our experiments with the SIMMC 2.0 \citep{kottur-etal-2021-simmc} dataset. Further, we  follow the experimental setup of the  SIMMC 2.0 challenge with the same dataset splits, inference time limitations, and performance metrics.
 
 \textbf{Implementation: } We conduct our experiments using PyTorch Huggingface’s transformers \citep{journals/corr/abs-1910-03771}.  All SimpleMTOD  model variants were initialized with  Open AI GPT-2 pretrained weights and  exhibits computational speed identical to Open AI GPT-2. We use Adam optimizer \citep{Adam}  with default parameter of Huggingface's AdamW implementation (\(lr=1e-3, eps= 1e-6, weight\_decay=0\)). 
 
 We use the GPT-2 tokenizer for encoding user and system utterances. However, we noticed that the default tokenizer encoder mechanism chunks special tokens introduced for visual object representation. Therefore, we implemented  an encoding mechanism which selectively skips the default byte-pair encoding for object tracking tokens.
 
 \textbf{Evaluation:} We use the same evaluation metrics and evaluation scripts provided with the SIMMC2.0 challenge. Table \ref{tbl:eval_met} shows metrics that are used for evaluating each benchmark task.
 
\section{Salience scores}
For the discussion we use input X gradient (IG) method from \cite{alammar-2021-ecco} as suggested in \cite{conf/emnlp/AtanasovaSLA20}. In the IG method of input saliency, attribution values are calculated across the embedding dimensions. With the values from embeddings dimension, the L2 norm is used to obtain a score per each token  Then resulting values are normalized by dividing by the sum of the attribution scores for all the tokens in the sequence.
Here we provide actual salience scores for heat-maps provided in the discussion in Section: \ref{sec:discussion}. 
\begin{figure*}[t!]
    \centering
    \includegraphics[width=2\columnwidth]{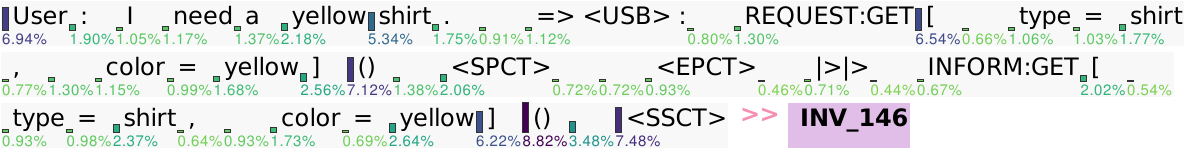}
    \captionsetup{justification=centering, width=2\columnwidth}
    \caption{Salience score when predicting the token \textit{INV\_146} for utterance \textit{I need a yellow shirt} without scene information.}
    \label{fig:ecco_coref_woscene_deteailed}
\end{figure*}
\vspace{1cm}
\begin{figure*}[t!]
    \centering
    \includegraphics[width=2\columnwidth]{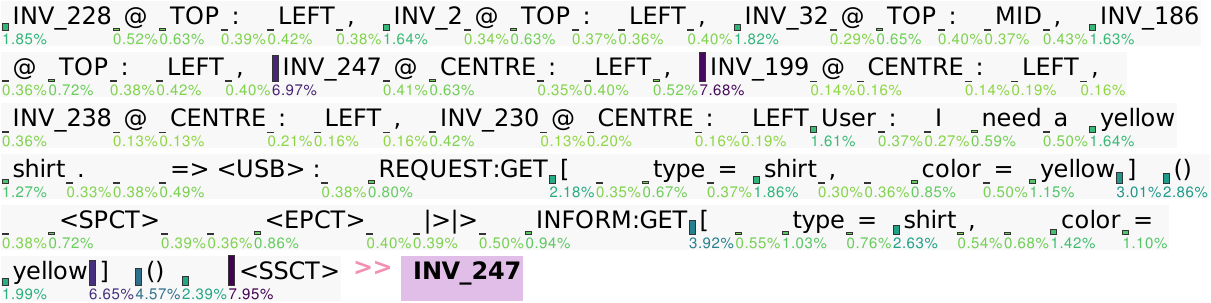}\label{sugfig:247_detailed}
    \captionsetup{justification=centering, width=2\columnwidth}
    \caption{Salience scores  {\it with scene information} when predicting the token \textit{INV\_247} in utterance \textit{I need a yellow shirt}.}
    \label{fig:ecco_coref_scene_detailed}
\end{figure*}
\vspace{1cm}
\begin{figure*}[t!]
    \centering
    \includegraphics[width=2\columnwidth]{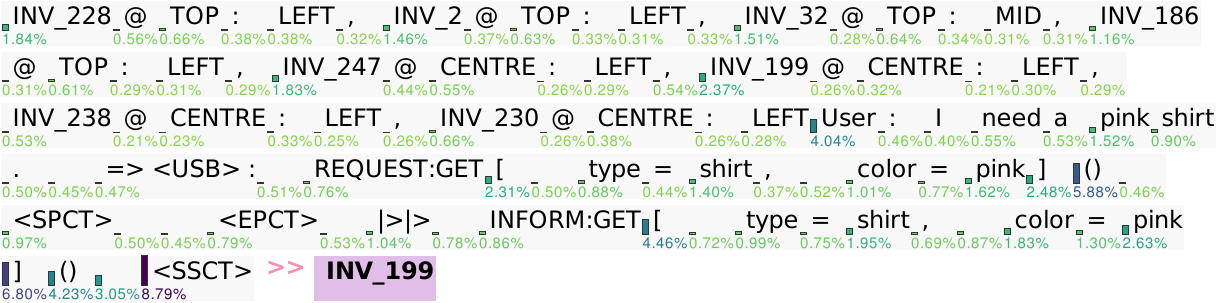}
    \captionsetup{justification=centering, width=2\columnwidth}
    \caption{Salience scores when predicting the token \textit{INV\_199}   for modified utterance \textit{I need a pink shirt} }
    \label{fig:ecco_coref_mod_detailed}
\end{figure*}
 \end{document}